\documentclass[9pt,twocolumn,twoside]{optica}
\setboolean{shortarticle}{false}
\setboolean{minireview}{false}

\usepackage{amsmath,amsfonts,amssymb,bbm}
\usepackage{graphicx}
\usepackage{bm}

\newcommand\measurementvec{\mathbf{b}} 
\newcommand\measurementmtx{\mathbf{H}}
\newcommand\imagevec{\mathbf{v}}
\newcommand{\crop}{\mathbf{C}}
\newcommand{\ADMMcrop}{\mathbf{D}}

\newcommand{\argmin}{\text{argmin }}

\title{DiffuserCam: Lensless Single-exposure 3D Imaging}

\author[1,*]{Nick Antipa}
\author[1,*]{Grace Kuo}
\author[1]{Reinhard Heckel}
\author[1]{Ben Mildenhall}
\author[1]{Emrah Bostan}
\author[1]{Ren Ng}
\author[1,+]{Laura Waller}

\affil[1]{Department of Electrical Engineering \& Computer Sciences, University of California, Berkeley, California 94720, USA}

\affil[*]{Authors contributed equally}

\affil[+]{Corresponding Author: lwaller@alum.mit.edu}

\dates{Compiled \today}

\ociscodes{(110.6880) Three-dimensional image acquisition; (110.1758) Computational imaging.}

\doi{\url{http://dx.doi.org/10.1364/optica.XX.XXXXXX}}

\begin{abstract}
We demonstrate a compact and easy-to-build computational camera for single-shot 3D imaging. Our lensless system consists solely of a diffuser placed in front of a standard image sensor. Every point within the volumetric field-of-view projects a unique pseudorandom pattern of caustics on the sensor. By using a physical approximation and simple calibration scheme, we solve the large-scale inverse problem in a computationally efficient way. The caustic patterns enable compressed sensing, which exploits sparsity in the sample to solve for more 3D voxels than pixels on the 2D sensor. Our 3D voxel grid is chosen to match the experimentally measured two-point optical resolution across the field-of-view, resulting in 100 million voxels being reconstructed from a single 1.3 megapixel image.  However, the effective resolution varies significantly with scene content. Because this effect is common to a wide range of computational cameras, we provide new theory for analyzing resolution in such systems.
\end{abstract}

\setboolean{displaycopyright}{true}
\begin{document}
\maketitle
\thispagestyle{fancy}
\ifthenelse{\boolean{shortarticle}}{\abscontent}{}



\section{Introduction} \label{intro}
\begin{figure*}
	\centering
    \includegraphics[width=0.9\linewidth]{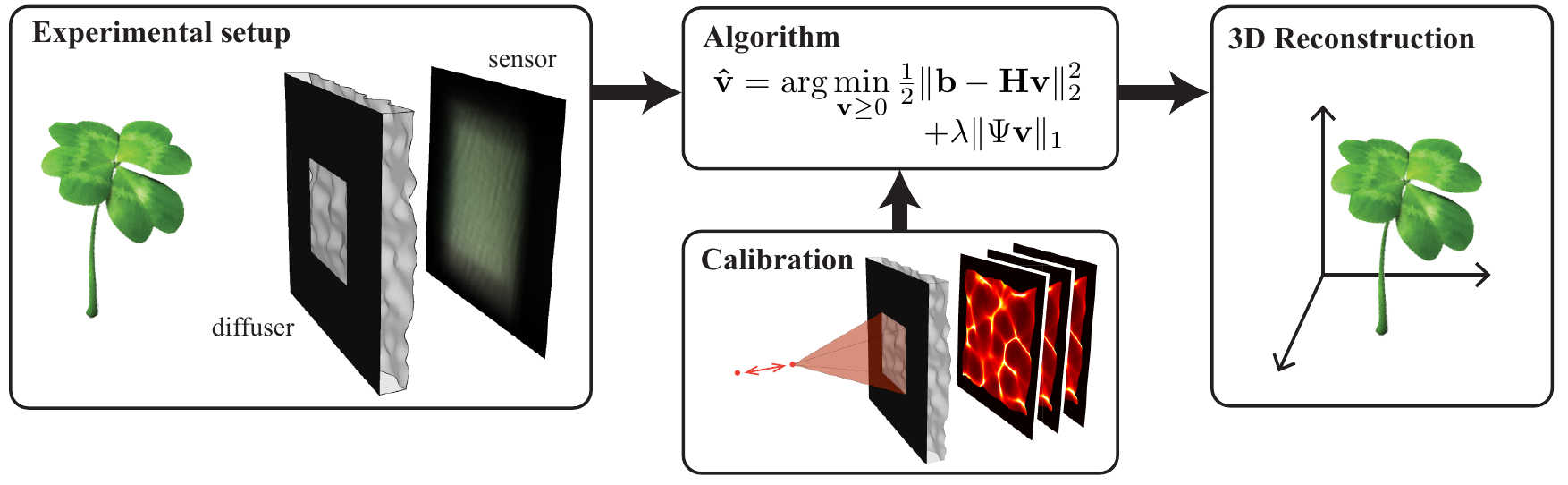}
	\caption{DiffuserCam setup and reconstruction pipeline. Our lensless system consists of a diffuser placed in front of a sensor (bumps on the diffuser are exaggerated for illustration). The system encodes a 3D scene into a 2D image on the sensor. A one-time calibration consists of scanning a point source axially while capturing images. Images are reconstructed computationally by solving a nonlinear inverse problem with a sparsity prior. The result is a 3D image reconstructed from a single 2D measurement.}
	\label{fig:setup}
\end{figure*}

Because optical sensors are 2D, capturing 3D information requires projection onto a 2D sensor in such a way that the 3D data can be recovered. Scanning and multi-shot methods can achieve high spatial resolution, but sacrifice capture speed and require complex hardware. In contrast, existing single-shot methods are fast but have low resolution. Most 3D imagers require bulky hardware, such as large bench-top microscopes. In this work, we introduce a compact and inexpensive single-shot lensless system capable of volumetric imaging, and show that it improves upon the sampling limit of existing single-shot systems by leveraging compressed sensing. 

We present DiffuserCam, a lensless imager that uses a diffuser to encode the 3D intensity of volumetric objects into a single 2D image. The diffuser, a thin phase mask, is placed a few millimeters in front of a traditional 2D image sensor. Each point source in 3D space creates a unique pseudorandom caustic pattern that covers a large portion of the sensor. Due to these properties, compressed sensing algorithms can be used to reconstruct more voxels than pixels captured, provided that the 3D sample is sparse in some domain. We recover this 3D information by solving a sparsity-constrained optimization problem, using a physical model and simple calibration scheme to make the computation and calibration scalable. This allows us to reconstruct on a grid of 100 million voxels, several orders of magnitude more than previous work. 

We demonstrate a prototype DiffuserCam system built entirely from commodity hardware. It is simple to calibrate, does not require precise alignment during construction and is light efficient (as compared to amplitude masks). We reconstruct 3D objects on a grid of 100 million voxels (non-uniformly spaced) from a single 1.3 megapixel image, which provides high resolution across a large volume. Our reconstructions show true depth sectioning, allowing us to generate 3D renderings of the sample.

Our system uses a nonlinear reconstruction algorithm, which results in object-dependent performance. We quantify this by experimentally measuring the resolution of our prototype using a variety of test targets, and we show that the standard two-point resolution criterion is misleading and should be considered a best-case scenario. Additionally, we propose a new local condition number analysis that explains the variable resolving power and show that the theory is consistent with our experiments.
 
DiffuserCam uses concepts from lensless camera technology and imaging through complex media, integrated together via computational imaging design principles. This new device could enable high-resolution lensless 3D imaging of large and dynamic 3D samples in an extremely compact package. Such cameras will open up new applications in remote diagnostics, mobile photography and \textit{in vivo} microscopy.

To review related previous work, we start with lensless cameras for 2D photography, which have shown great promise because of their small form factors. Unlike traditional cameras, in which a point in the scene maps to a pixel on the sensor, lensless cameras map a point in the scene to many points on the sensor, requiring computational reconstruction. A typical lensless architecture replaces the lens with an encoding element placed directly in front of the sensor. Incoherent 2D lensless cameras have been demonstrated using amplitude masks~\cite{Rice_FlatCam}, diffractive masks~\cite{RAMBUS,gill2017thermal}, random reflective surfaces~\cite{random_lenses,stylianou2016sparklegeometry}, and modified microlens arrays~\cite{TOMB0_tanida2001thin} as the encoding element.  2D lensless imaging with coherent illumination has been demonstrated in \cite{harm2014lensless, chi2011optical}, and extended to 3D~\cite{Brady_compressive_holography:09,lee2016exploiting_speckle_3D_lensless,bishara2010lensfree,faulkner2004movable_ptychography}, but these methods require active or coherent illumination. Our system uses a similar lensless architecture but extends both the design and image reconstruction to enable 3D capture without the need for coherent lighting.

Light field cameras, also called integral imagers, passively capture 4D space-angle information in a single-shot~\cite{Ng_handheld}, which can be used for 3D reconstructions. This concept can be built into a thin form factor with microlens arrays~\cite{TOMBO_3d_Horisaki2007} or Fresnel zone plates~\cite{Tajima_iccp2017_lenslessLF}. Lenslet array-based 3D capture schemes have also been used in microscopy~\cite{Levoy:2006:LightFieldMicroscopy}, where improved results are obtained when wave-optical effects~\cite{Broxton:13, Herbert_Liu_ps_scatter:15} or \textit{in vivo} tissue scattering~\cite{pegard2016compressive, Herbert_Liu_ps_scatter:15} are accounted for. All of these systems must trade resolution or field-of-view for single-shot capture. DiffuserCam, in contrast, uses compressed sensing to capture large 3D volumes at high resolution in a single exposure.

Diffusers are often used as a proxy for general scattering media in the context of developing methods for imaging through scattering~\cite{katz2014non, edrei2016memory, singh2014looking}. These works have similar mathematical models to our system, but instead of trying to mitigate the effects of unwanted scattering, here we use the diffuser as an optical element in our system design. Therefore, we choose a thin, optically smooth diffuser that refracts pseudorandomly (as opposed to true random scattering). Such diffusers have been shown to produce high contrast patterns under incoherent illumination, enabling light field imaging ~\cite{Antipa_iccp2016_diffuserLF}, and have also been used to record coherent holograms ~\cite{lee2016exploiting_speckle_3D_lensless, Kashter:17}. Multiple scattering with coherent illumination has been demonstrated as an encoding mechanism for 2D compressed sensing~\cite{liutkus2014_CS_scattering}, but necessitates an inefficient transmission matrix calibration approach, limiting reconstructions to a few thousand pixels. We achieve similar benefits without needing coherent illumination, and, unlike previous work, we use compressed sensing to add depth information. Finally, our system is designed to enable simpler calibration and more efficient computation, allowing for 3D reconstruction at megavoxel scales with superior image quality.

\subsection{System Overview}\label{sec:system}

DiffuserCam is part of the class of mask-based lensless imagers in which a phase or amplitude mask is placed a small distance in front of a sensor, with no main lens. Our mask (the diffuser) is a thin transparent phase object with smoothly varying thickness (see Fig.~\ref{fig:setup}). When illuminated by an incoherent source sufficiently far from the sensor, the convex bumps concentrate light into high-frequency pseudorandom caustic patterns which are captured by the sensor. The caustic patterns, termed Point Spread Functions (PSFs), vary with the 3D position of the source, thereby capturing 3D information.

\begin{figure}[t]
	\center
    \includegraphics[width=\linewidth]{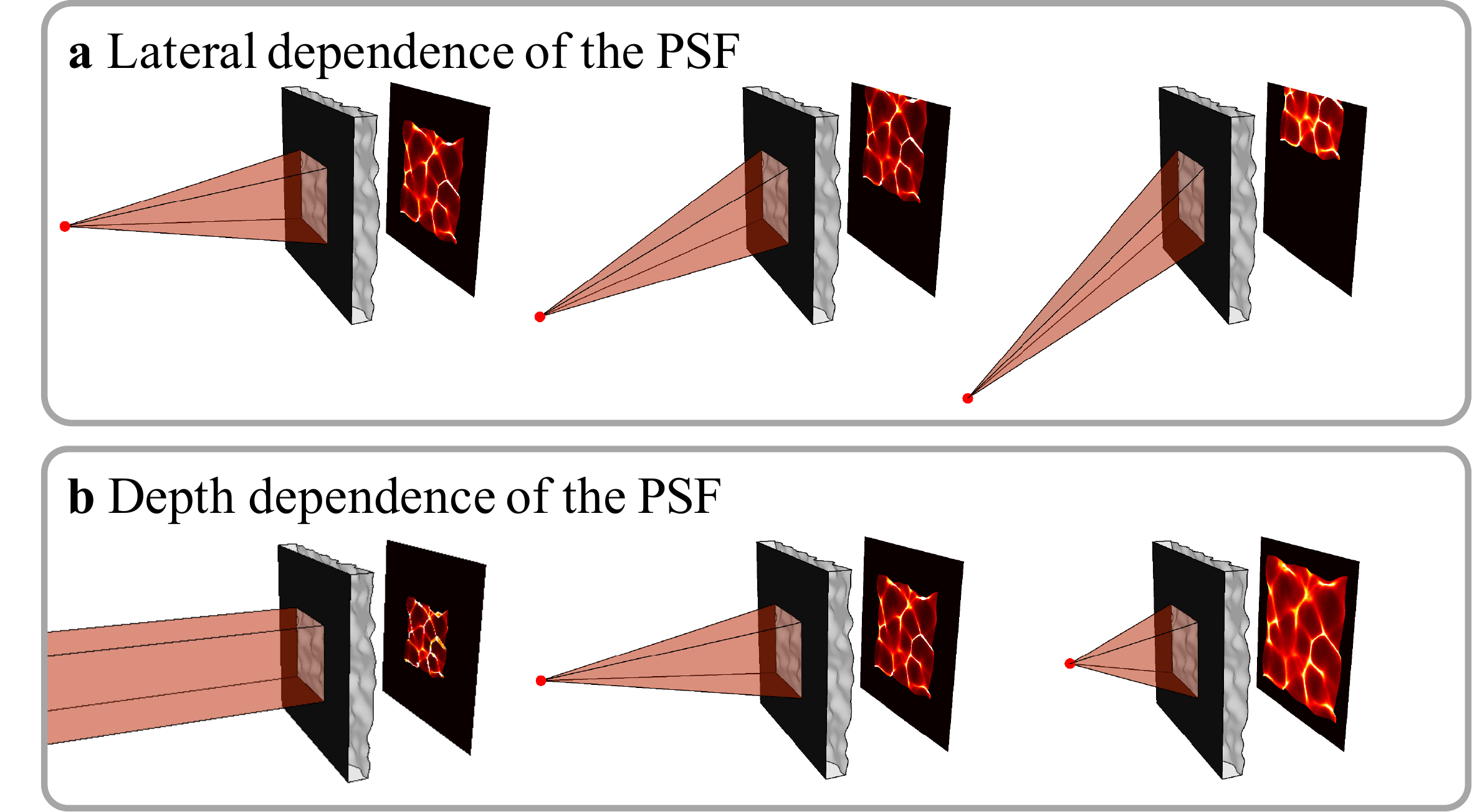}
	\caption{Caustic patterns shift with lateral shifts of a point source in the scene and scale with axial shifts. (a) Ray-traced renderings of caustics as the point source moves laterally. For large shifts (far right), part of the pattern is clipped by the sensor. (b) The caustics approximately magnify as the source is brought closer. }
	\label{fig:psfShift}
\end{figure}

To illustrate how the caustics encode 3D information, Fig.~\ref{fig:psfShift} shows simulations of caustic PSFs as a function of point source location in object space. A lateral shift of the point source causes a lateral translation of the PSF. An axial shift of the point source causes (approximately) a scaling of the PSF. Hence, each 3D position in the volume generates a unique PSF. The resolution of our camera depends on the structure and spatial frequencies present in the caustic patterns. Because the caustics retain high spatial frequencies over a large range of depths, DiffuserCam attains good lateral resolution for objects at any depth within the volumetric field-of-view (FoV). 

By assuming that all points in the scene are incoherent with each other, the measurement can be modeled as a linear combination of PSFs from different 3D positions. We represent this as matrix-vector multiplication: 
\begin{equation} \label{eq:linear_system}
    \measurementvec = \measurementmtx \imagevec,
\end{equation}
where $\measurementvec$ is a vector representing the 2D sensor measurement and $\imagevec$ is a vector representing the intensity of the object at every point in the 3D FoV, sampled on a user-chosen grid (discussed in Section \ref{analysis}). $\measurementmtx$ is the forward model matrix whose columns consist of each of the caustic patterns created by the corresponding 3D points on the object grid. The number of entries in $\measurementvec$ and the number of rows of $\measurementmtx$ are equal to the number of pixels on the image sensor, but the number of columns in $\measurementmtx$ is set by the choice of reconstruction grid. Note that this model does not account for partial occlusion of sources.

In order to reconstruct the 3D image, $\imagevec$, from the measured 2D image, $\measurementvec$, we must solve~\eqref{eq:linear_system} for $\imagevec$. However, if we use the full optical resolution of our system (see Sec. \ref{analysis}\ref{resolution}), $\imagevec$ will exist on a grid that contains more voxels than there are pixels on the sensor. In this case, $\measurementmtx$ has many more columns than rows, so the problem is underdetermined and we cannot uniquely recover $\imagevec$ simply by inverting~\eqref{eq:linear_system}. To remedy this, we rely on sparsity-based computation principles~\cite{candes2008_Compressed_Sensing}. We exploit the fact that the object can be represented in a domain in which it is sparse (few non-zero elements) and solve the $\ell_1$ regularized inverse problem:

\begin{equation}  
\label{eq:basis_pursuit}
\hat{\imagevec} = \underset{\imagevec\geq 0}{\argmin}\tfrac{1}{2}\|\measurementvec - \measurementmtx \imagevec\|_2^2 + \lambda \|\Psi\imagevec\|_1.
\end{equation} 

\noindent Here, $\Psi$ is a linear transform that sparsifies the 3D object, $\imagevec\geq 0$ is a nonnegativity constraint, and $\lambda$ is a tuning parameter. For objects that are sparse in native 3D space, such as fluorescent samples, we choose $\Psi$ to be the identity matrix. For objects that are not natively sparse, but whose gradient is sparse, we choose $\Psi$ to be the finite difference operator, so $\|\Psi \imagevec\|_1$ is the 3D Total Variation (TV) semi-norm~\cite{rudin1992_total_variation_TV_denoise}. 

Equation~(\ref{eq:basis_pursuit}) is the basis pursuit problem in compressed sensing~\cite{candes2008_Compressed_Sensing}. For this optimization procedure to succeed, $\measurementmtx$ must have distributed, uncorrelated columns. The diffuser creates caustics that spread across many pixels in a pseudorandom fashion and contain high spatial frequencies. Therefore, any shift or magnification of the caustics leads to a new pattern that is uncorrelated with the original one. As discussed in Sections \ref{methods}\ref{compRecon} and \ref{methods}\ref{sec:inverse}, these two properties lead to a matrix that allows us to reconstruct 3D images $\imagevec$ from the measurement $\measurementvec = \measurementmtx \imagevec$.

\section{Methods} \label{methods}

\subsection{System Architecture} \label{sec:architecture}
\begin{figure*}
	\centering
    \includegraphics[width=.95\linewidth]{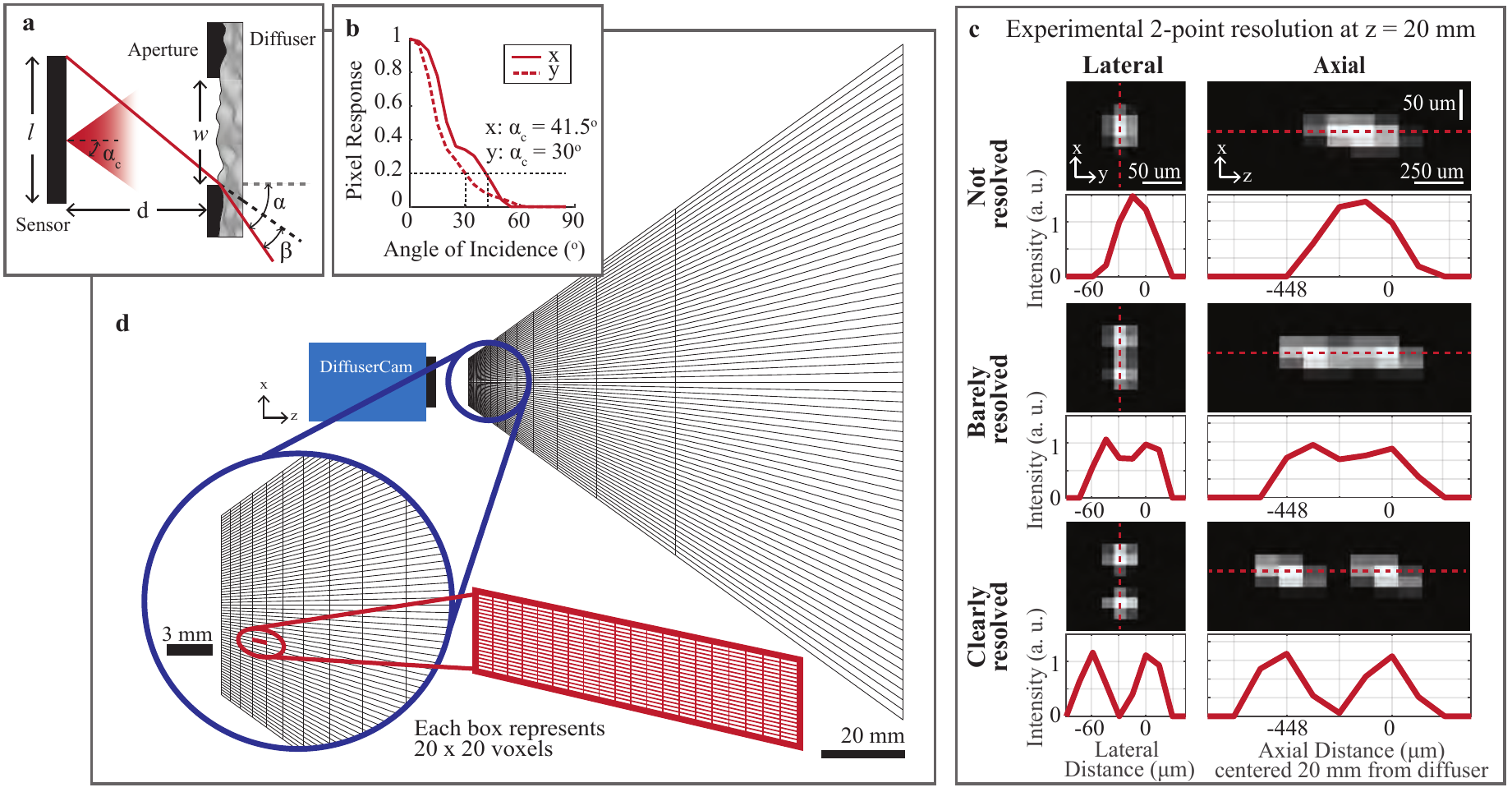}
	\caption{Experimentally determined field-of-view (FoV) and resolution. (a) System architecture with design parameters. (b) Angular pixel response of our sensor. We define the angular cutoff ($\alpha_c$) as the angle at which the response falls to 20\%. (c) Reconstructed images of two points (captured separately) at varying separations laterally and axially, near the z = 20 mm depth plane. Points are considered resolved if they are separated by a dip of at least 20\%. (d) To-scale non-uniform voxel grid for 3D reconstruction, viewed from above. The voxel grid is based on the system geometry and Nyquist-sampled two-point resolution over the entire FoV. For visualization purposes, each box represents 20$\times$20 voxels, as shown in red. }
	\label{fig:res_map}
\end{figure*}

The hardware setup for DiffuserCam consists of a diffuser placed at a fixed distance in front a sensor (see Fig.~\ref{fig:res_map}a). The convex bumps on the diffuser surface can be thought of as a set of randomly-spaced microlenses that have statistically varying focal lengths and f-numbers. The f-number determines the minimum feature size of the caustics, which sets our optical resolution. The average focal length determines the distance at which the caustics have highest contrast (the \textit{caustic plane}), which is where we place the sensor~\cite{Antipa_iccp2016_diffuserLF}. 

Our prototype is built using a PCO.edge 5.5 Color camera ($6.5 \mu m$ pixels). The diffuser is an off-the-shelf engineered diffuser (Luminit $0.5^\circ$) with a flat input surface and an output surface that is described statistically as Gaussian lowpass filtered white noise with an average spatial feature size of 140$\mu$m and average slope magnitude of $0.7^\circ$ (details in Supplementary Fig. S1). This corresponds to an average focal length of approximately $8$ mm and average f-number of $50$. Due to the high f-number, the caustics maintain high contrast over a large range of propagation distances. Therefore, the diffuser need not be placed precisely at the caustic plane; in our prototype, $d=8.9$ mm from our sensor. Additionally, we affix a 5.5 $\times$ 7.5 mm aperture directly on the textured side of the diffuser to limit the support of the caustics.

Similar to a traditional camera, the pixel pitch should Nyquist sample the minimum features of the caustics. The smallest features generated by the caustic patterns are roughly twice the pixel pitch on our sensor, so we perform 2x2 binning on the data, yielding 1.3 megapixel images, before applying our 3D reconstruction algorithm.

\subsection{Convolutional Forward Model} \label{compRecon}

Recovering a 3D image requires knowing the system transmission matrix, $\measurementmtx$, which is extremely large. Measuring or storing the full $\measurementmtx$ would be impractical, requiring millions of calibration images and operating on multi-Terabyte matrices. The convolution model outlined below drastically reduces complexity of both calibration and computation. 

We describe the object, $\imagevec$, as a set of point sources located at $(x,y,z)$ on a non-Cartesian grid and represent the relative radiant power collected by the aperture from each source as $\imagevec(x,y,z)$. The caustic pattern at pixel $(x',y')$ on the sensor due to a unit-powered point source at $(x,y,z)$ is the PSF, $h(x',y'; x,y,z)$. Thus, $\measurementvec(x',y')$ represents the 2D sensor measurement recorded after the light from every point in $\imagevec$ propagates through the diffuser and onto the sensor. This lets us explicitly write the matrix-vector multiplication $\measurementmtx \imagevec$ by summing over all voxels in the FoV:

\begin{equation} \label{eq:sum_form}
    \measurementvec(x',y') = \sum_{(x,y,z)}\imagevec(x,y,z)h(x',y';x,y,z).
\end{equation}

The convolution model amounts to a shift invariance (or infinite memory effect~\cite{katz2014non, edrei2016memory}) assumption, which greatly simplifies the evaluation of~\eqref{eq:sum_form}. Consider the caustics created by point sources at a fixed distance, $z$, from the diffuser. Because the diffuser surface is slowly varying and smooth, the paraxial approximation holds. This implies that a lateral translation of the source by $(\Delta x,\Delta y)$ leads to a lateral shift of the caustics on the sensor by $(\Delta x',\Delta y') = (m \Delta x,m\Delta y)$, where $m$ is the paraxial magnification. We validate this behavior in both simulations (see Fig.~\ref{fig:psfShift}) and experiments (see Section \ref{analysis}\ref{sec:mismatch}). For notational convenience, we define the on-axis caustic pattern at depth $z$ as $h(x',y';z) \mathrel{\mathop:}= h(x',y';0,0,z)$. Thus, the off-axis caustic pattern is given by $h(x',y'; x, y, z)=h(x'+m x,y'+ m y;z)$. Plugging into~\eqref{eq:sum_form}, the sensor measurement is then given by:
\begin{equation} \label{eq:convolution}
	\begin{aligned}
	    \measurementvec(x',y') & = \sum_z\sum_{(x,y)}\imagevec(x,y,z)h(x'+m x,y'+m y;z)\\
	    &=\crop\sum_z\left[ \imagevec\left(\frac{-x'}{m},\frac{-y'}{m},z\right)*h\left(x',y';z\right)\right].
    \end{aligned}
\end{equation}
Here, $*$ represents 2D discrete convolution over $(x',y')$, which returns arrays that are larger than the originals. Hence, we crop to the original sensor size, denoted by the linear operator $\crop$. For an object discretized into $N_z$ depth slices, the number of columns of $\measurementmtx$ is $N_z$ times larger than the number of elements in $\measurementvec$ (\textit{i.e.} the number of sensor pixels), so our system is underdetermined.

The cropped convolution model provides three benefits. First, it allows us to compute $\measurementmtx \imagevec$ as a linear operator in terms of $N_z$ images, rather than instantiating $\measurementmtx$ explicitly (which would require petabytes of memory to store). In practice, we evaluate the sum of 2D cropped convolutions using a single circular 3D convolution, enabling the use of 3D FFTs, which scale well to large array sizes (for details, see Supplementary Sec. 2). Second, it provides us with a theoretical justification of our system's capability for sparse reconstruction. Derivations in \cite{krahmer_suprema_2014} show that translated copies of a random pattern provide close-to-optimal compressed sensing performance. 

The third benefit of our convolution model is that it enables simple calibration. Rather than measuring the system response from every voxel (millions of images per depth), we only need to capture a single calibration image of the caustic pattern from an on-axis point source at each depth plane.\footnote{While the scaling effect described in Sec.~\ref{intro}\ref{sec:system} suggests that we could use only one image for calibration and scale it to predict PSFs at different depths, we find that there are subtle changes in the caustic structure with depth. Hence, we obtain better results when we calibrate with PSFs measured at each depth.} A typical calibration thus consists of capturing images as a single point source is moved axially. This takes minutes, but need only be performed once. The added aperture on DiffuserCam ensures that a point source at the minimum $z$ distance generates caustics that just fill the sensor, so that the entire PSF is captured in each image (see Supplementary Fig. S2).


\subsection{Inverse Algorithm} \label{sec:inverse}
Our inverse problem is extremely large in scale, with millions of inputs and outputs. Even with the convolution model described above, using projected gradient techniques is extremely slow due to the time required to compute the proximal operator of 3D total variation \cite{beck2009fast_TV}. To alleviate this, we use the Alternating Direction Method of Multipliers (ADMM)~\cite{Boyd.etal2011}. We derive a variable splitting that leverages the specific structure of our problem.
 
Our algorithm uses the fact that $\Psi$ can be written as a circular convolution for both the 3D TV and native sparsity cases. Additionally, we factor the forward model in~\eqref{eq:convolution} into a diagonal component, $\ADMMcrop$, and a 3D convolution matrix, $\mathbf{M}$, such that $\measurementmtx = \ADMMcrop \mathbf{M}$ (details in Sec. ). Thus, both the forward operator and the regularizer can be computed in 3D Fouier space. This enables us to use variable-splitting~\cite{Almeida.Figueiredo2013,Matakos.etal2013,Afonso_var_split_admm.etal2010} to formulate the constrained counterpart of~\eqref{eq:basis_pursuit}:

\begin{equation}\label{eq:basis_persuit_constraint} \begin{aligned} \hat{\imagevec} = \underset{w\geq 0,u,v}{\argmin}&\tfrac{1}{2}\|\measurementvec - \ADMMcrop v\|_2^2 + \lambda \|u\|_1 \\ \mbox{s.t. } &v = \mathbf{M} \imagevec, u = \Psi \imagevec, w = \imagevec\text{,} \end{aligned} \end{equation} 

\noindent where $v,u$, and $w$ are auxiliary variables. We solve~\eqref{eq:basis_persuit_constraint} by following the commonly-used augmented Lagrangian arguments~\cite{Nocedal.Wright2006}. Using ADMM, this results in the following scheme at iteration $k$: 
\begin{align*} &u^{k+1} \gets \mathcal{T}_{\frac{\lambda}{\mu_2}}\left(\Psi \imagevec^k + \eta^k \big/ \mu_2\right)\\ &v^{k+1} \gets (\ADMMcrop^\intercal \ADMMcrop + \mu_1 I)^{-1}\left(\xi^k + \mu_1 \mathbf{M}\imagevec^k + \ADMMcrop^\intercal \measurementvec\right)\\ &w^{k+1} \gets \max\left(\rho^k \big/ \mu_3+\imagevec^k,0\right)\\ &\imagevec^{k+1} \gets \left(\mu_1 \mathbf{M}^\intercal \mathbf{M} + \mu_2 \Psi^\intercal \Psi + \mu_3 I\right)^{-1}r^{k}\\ &\xi^{k+1} \gets \xi^k + \mu_1(\mathbf{M}\imagevec^{k+1}-v^{k+1})\\ &\eta^{k+1} \gets \eta^k + \mu_2(\Psi \imagevec^{k+1} - u^{k+1})\\ &\rho^{k+1} \gets \rho^k + \mu_3(\imagevec^{k+1} - w^{k+1})\text{,} \end{align*} 

\noindent where 
$$r^{k} = (\mu_3 w^{k+1}-\rho^k) + \Psi^\intercal\left(\mu_2u^{k+1}-\eta^k\right)+ \mathbf{M}^\intercal\left(\mu_1v^{k+1}- \xi^k \right)\hspace{-0.23em}.$$ 

Note that $\mathcal{T}_{\nu}$ is a vectorial soft-thresholding operator with a threshold value of $\nu$ \cite{Wang.etal2008}, and $\xi$, $\eta$ and $\rho$ are the Lagrange multipliers associated with $v$, $u$, and $w$, respectively. The scalars $\mu_1$, $\mu_2$ and $\mu_3$ are penalty parameters which we compute automatically using the tuning strategy in~\cite{Boyd.etal2011}.

Although our algorithm involves two large-scale matrix inversions, both can be computed efficiently and in closed form. Since $\ADMMcrop$ is diagonal, $(\ADMMcrop^\intercal \ADMMcrop + \mu_1 I)$ is itself diagonal, requiring complexity $\mathcal{O}(n)$ to invert using point-wise multiplication. Additionally, all three matrices in $(\mu_1 \mathbf{M}^\intercal \mathbf{M} + \mu_2 \Psi^\intercal \Psi + \mu_3 I)$ are diagonalized by the 3D discrete Fourier transform (DFT) matrix, so inversion of the entire term can be done using point-wise division in 3D frequency space. Therefore, its inversion has good computational complexity, $\mathcal{O}(n^3 \log n)$, since it is dominated by two 3D FFTs being applied to $n^3$ total voxels. We parallelize our algorithm on the CPU using C++ and Halide~\cite{ragan2013halide}, a high performance programming language specialized for image processing (A comparison of regularizers and runtimes is shown in Supplementary Sec. 2c).

A typical reconstruction requires at least 200 iterations. Solving for $2048\times2048\times128=537$ million voxels takes 26 minutes (8 seconds per iteration) on a 144-core workstation and requires 85 Gigabytes of RAM. A smaller reconstruction ($512\times512\times128=33.5$ million voxels) takes 3 minutes (1 second per iteration) on a 4-core laptop with 16 Gigabytes of RAM.

\section{System Analysis} \label{analysis}

Unlike traditional cameras, the performance of computational cameras depends on properties of the scene being imaged (\textit{e.g.} the number of sources). As a consequence, standard two-point resolution metrics may be misleading, as they do not predict resolving power for more complex objects.  To address this, we propose a new local condition number metric that better predicts performance. We analyze resolution, FoV and the validity of the convolution model, then combine these analyses to determine the appropriate sampling grid for faithfully reconstructing real-world objects.


\subsection{Field-of-View} \label{sec:fov}

\begin{figure}[t!]
	\center
    \includegraphics[width=\linewidth]{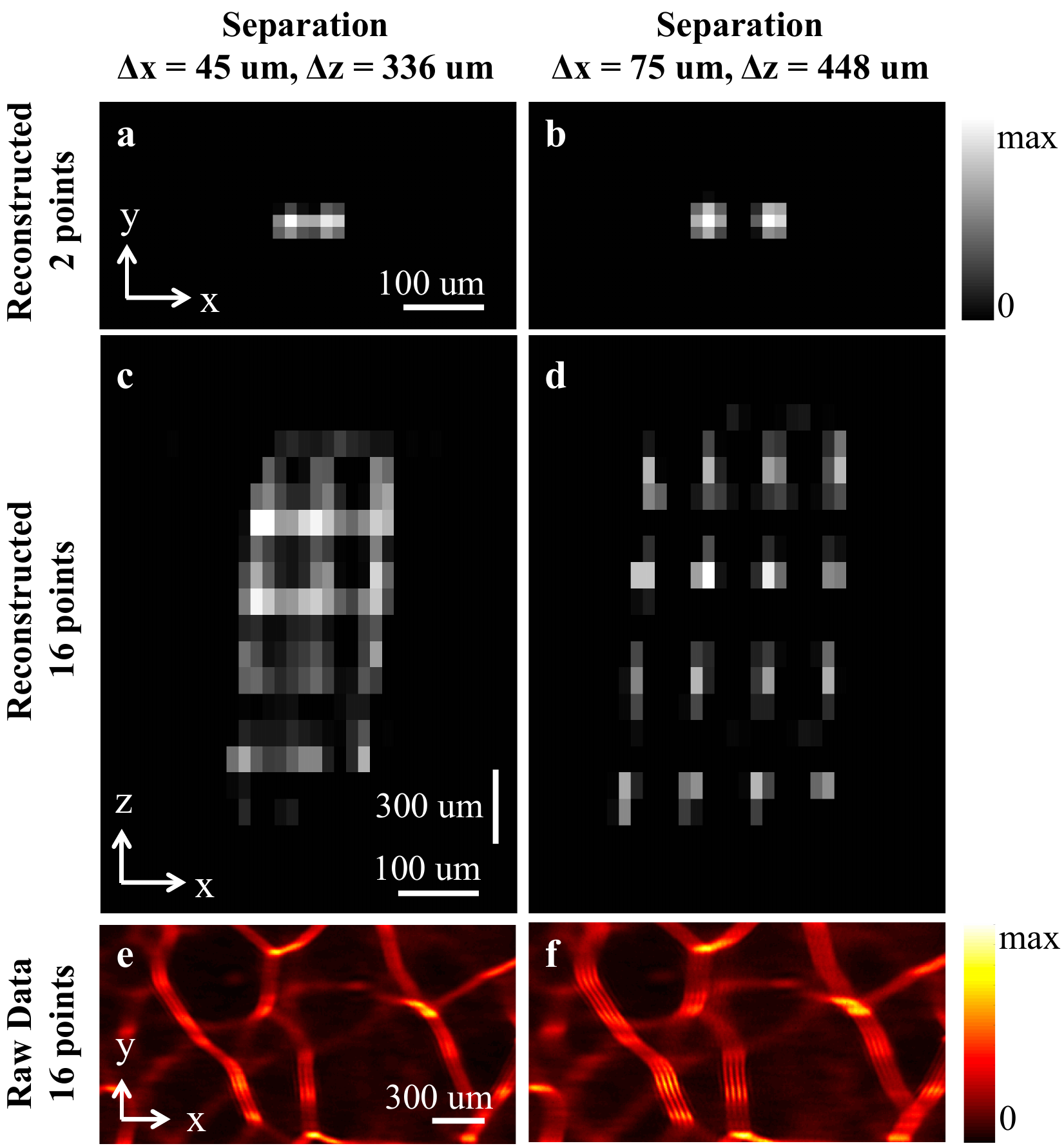}
	\caption{Our computational camera has object-dependent performance, such that the resolution depends on the number of points. (a) To illustrate, we show here a situation with two points successfully resolved at the two-point resolution limit $(\Delta x, \Delta z)=(45 \mu m, 336 \mu m)$ at a depth of approximately 20 mm. (c) However, when the object consists of more points (16 points in a 4$\times$4 grid in the $x-z$ plane) at the same spacing, the reconstruction fails. (b,d) Increasing the separation to $(\Delta x, \Delta z)=(75 \mu m, 448 \mu m)$ gives successful reconstructions. (e,f) A close-up of the raw data shows noticeable splitting of the caustic lines for the 16 point case, making the points distinguishable. Heuristically, the 16 point resolution cutoff is a good indicator of resolution for real-world objects.}
	\label{fig:16point}
\end{figure}

At every depth in the volume, the angular half-FoV is determined by the most extreme lateral position that contributes to the measurement. There are two possible limiting factors. The first is the geometric angular cutoff, $\alpha$, set by the aperture size, $w$, the sensor size, $l$, and the distance from the diffuser to the sensor, $d$ (see Fig.~\ref{fig:res_map}a). Since the diffuser bends light, we also take into account the diffuser's maximum deflection angle, $\beta$. This gives a geometric angular half-FoV at every depth of $l+w=2d\tan (\alpha -\beta)$. The second limiting factor is the angular response of the sensor pixels. Real-world sensor pixels may not accept light at the high angles of incidence that our lensless camera accepts, so the sensor angular response (shown in Fig.~\ref{fig:res_map}b) may limit the FoV. Defining the angular cutoff of the sensor, $\alpha_c$, as the angle at which the camera response falls to 20\% of its on-axis value, we can write the overall FoV equation as:

\begin{equation}  \label{eq:FOV}
\text{FoV} = \beta + \min[\alpha_c, \tan^{-1}(\tfrac{l+w}{2d})].
\end{equation}

Since we image in 3D, we must also consider the axial FoV. In practice, the axial FoV is limited by the range of calibrated depths. However, the system geometry creates bounds on possible calibration locations. Calibration cannot be arbitrarily close to the sensor since the caustics would exceed the sensor size. To account for this, we impose a minimum object distance such that an on-axis point source creates caustics that fill the sensor. Theoretically, our system can capture sources infinitely far away, but the axial resolution degrades with depth. The hyperfocal plane represents the axial distance after which no more axial resolution is available, establishing an upper bound. Objects beyond the hyperfocal focal plane have no depth information, but can be reconstructed to create 2D images for photographic applications~\cite{kuo2017diffusercam}, without any hardware modifications.

In our prototype, the axial FoV ranges from the minimum calibration distance (7.3 mm) to the hyperfocal plane (2.3 m). The angular FoV is limited by the pixel angular acceptance ($\alpha_c = 41.5^\circ$ in $x$, $\alpha_c = 30^\circ$ in $y$). Combined with our diffuser's maximum deflection angle ($\beta = 0.5^\circ$) this yields an angular FoV of $\pm42^\circ$ in $x$ and $\pm30.5^\circ$ in $y$. We validate the FoV experimentally by capturing a scene at optical infinity and measuring the angular extent of the result (see Supplementary Fig. S3).


\subsection{Resolution} \label{resolution}
Investigating optical resolution is critical for both quantifying system performance and choosing our reconstruction grid. Although the raw data is collected on a fixed sensor grid, we can choose the non-uniform 3D reconstruction grid arbitrarily. The choice of reconstruction grid is important. When the grid is chosen with voxels that are too large, resolution is lost, and when they are too small, extra computation is performed without resolution gain. In this section we explain how to choose the grid of voxels for our reconstructions, with the aim of Nyquist sampling the two-point optical resolution limit.

\subsubsection{Two-point resolution}

A common metric for resolution analysis in traditional cameras is two-point distinguishablity. We measure our system's two-point resolution by imaging scenes containing two point sources at different separation distances, built by summing together images of a single point source (1$\mu$m pinhole, wavelength $532 nm$) at two different locations. 

We reconstruct the scene using our algorithm, with $\lambda=0$ to remove the influence of the regularizer. To ensure best-case resolution, we use the full 5 MP sensor data (no $2\times 2$ binning). The point sources are considered distinguishable if the reconstruction has a dip of at least $20\%$ between the sources, as in the Rayleigh criterion. Figure~\ref{fig:res_map}c shows reconstructions with point sources separated both laterally and axially. 

Our system has highly non-isotropic resolution (Fig.~\ref{fig:res_map}d), but we can use our model to predict the two-point distinguishability over the entire volume from localized experiments. Due to the shift invariance assumption, the lateral resolution is constant within a single depth plane and the paraxial magnification causes the lateral resolution to vary linearly with depth. For axial resolution, the main difference between two point sources is the size of their PSF supports. We find pairs of depths such that the difference in their support widths is constant:
\begin{equation}  \label{eq:zres}
c = \tfrac{1}{z_1} - \tfrac{1}{z_2}.
\end{equation}
Here, $z_1$ and $z_2$ are neighboring depths and $c$ is a constant determined experimentally. 

Based on this model, we set the voxel spacing in our grid to Nyquist sample the 3D two-point resolution. Figure \ref{fig:res_map}d shows a to-scale map of the resulting voxel grid. Axial resolution degrades with distance until it reaches the hyperfocal plane ($\sim$2.3 m from the camera), beyond which no depth information is recoverable. Due to the non-telecentric nature of the system, the voxel sizes are a function of depth, with the densest sampling occurring close to the camera. Objects within 5 cm of the camera can be reconstructed with somewhat isotropic resolution; this is where we place objects in practice.

\subsubsection{Multi-point resolution}

In a traditional camera, resolution is a function of the system and is independent of the scene. In contrast, computational cameras that use nonlinear reconstruction algorithms may incur degradation of the effective resolution as the scene complexity increases. To demonstrate this in our system, we consider a more complex scene consisting of 16 point sources. Figure~\ref{fig:16point} shows experiments using 16 point sources arranged in a 4$\times$4 grid in the $(x,z)$ plane at two different spacings. The first spacing is set to match the measured two-point resolution limit ($\Delta x$=45$\mu$m, $\Delta z$=336$\mu$m). Despite being able to separate two points at this spacing, we cannot resolve all 16 sources. However, if we increase the source separation to ($\Delta x$=75$\mu$m, $\Delta z$=448$\mu$m), all 16 points are distinguishable (Fig.~\ref{fig:16point}d). In this example, the usable lateral resolution of the system degrades by approximately $1.7\times$ due to the increased scene complexity. As we show in Section \ref{analysis}\ref{sec:conditioning}, the resolution loss does not become arbitrarily worse as the scene complexity increases.

This experiment demonstrates that existing resolution metrics cannot be blindly used to determine performance of computational cameras like ours. How can we then analyze resolution if it depends on object properties? In the next section, we introduce a general theoretical framework for assessing resolution in computational cameras.


\subsection{Local condition number theory} \label{sec:conditioning}
Our goal is to provide new theory  that describes how the effective reconstruction resolution of computational cameras changes with object complexity. To do so, we introduce a numerical analysis of our forward model to determine how well it can be inverted. 

First, note that recovering the image $\imagevec$ from the measurement $\measurementvec=\measurementmtx \imagevec$ comprises simultaneous estimation of the locations of all nonzeros within our image reconstruction, $\imagevec$, as well as the values at each nonzero location. To simplify the problem, suppose an oracle tells us the exact locations of every source within the 3D scene. This corresponds to knowing \textit{a priori} the support of $\imagevec$, so we then need only determine the \emph{values} of the nonzero elements in $\imagevec$. This can be accomplished by solving a least squares problem using a sub-matrix consisting of only the columns of $\measurementmtx$ that correspond to the indices of the nonzero voxels. If this problem fails, then the more difficult problem of simultaneously determining the nonzero locations \textit{and} their values will certainly fail. 

In practice, the measurement is corrupted by noise. 
The maximal effect this noise can have on the least-squares estimate of the nonzero values is determined by the condition number of the sub-matrix described above. We therefore say that the reconstruction problem is ill-posed if any sub-matrices of $\measurementmtx$ are very ill-conditioned. In practice, ill-conditioned matrices result in increased noise sensitivity and longer reconstruction times, as more iterations are needed to converge to a solution.

\begin{figure}
    \centering
    \includegraphics[width=\linewidth]{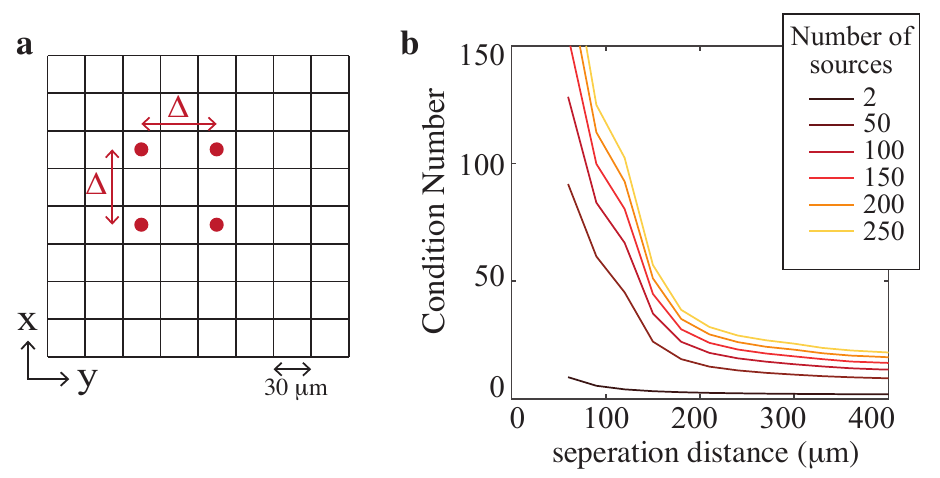}
    \caption{
Our local condition number theory shows how resolution varies with object complexity. (a) Virtual point sources are simulated on a fixed grid and moved by integer numbers of voxels to change the separation distance. (b) Local condition numbers are plotted for sub-matrices  corresponding to grids of neighboring point sources with varying separation (at depth 20 mm from the sensor). As the number of sources increases, the condition number approaches a limit, indicating that resolution for complex objects can be approximated by a limited number (but more than two) sources.}
\label{fig:condnr}
\end{figure}

The worst-case scenario in our system is when multiple sources are in a contiguous block, since nearby measurements are always most similar. Therefore, we compute the condition number of sub-matrices of $\measurementmtx$ corresponding to a group of point sources with separation varying by integer numbers of voxels. We repeat this calculation for different numbers of sources. The results are shown in Fig.~\ref{fig:condnr}. As expected, the conditioning is worse when sources are closer together. In this case, increased noise sensitivity means that even small amounts of noise could prevent us from resolving the sources. This trend matches what we saw experimentally in Figs.~\ref{fig:res_map} and~\ref{fig:16point}. 

Figure~\ref{fig:condnr} shows that the local condition number increases with the number of sources in the scene, as expected. This means that resolution will degrade as more and more sources are added.  We can see in Fig.~\ref{fig:condnr} that, as the number of sources increases linearly, the conditioning approaches a limiting case. Hence, the resolution does not become arbitrarily worse with increased number of sources. Therefore we can estimate the system resolution for complex objects from distiguishability measurements of a limited number of point sources. This is experimentally validated in Section \ref{sec:results}, where we find that the experimental 16-point resolution is a good predictor of the resolution for a USAF target.

Unlike traditional resolution metrics, our new local condition number theory explains the resolution loss we observe experimentally. We believe that it is sufficiently general to be applicable to other computational cameras, which likely exhibit similar performance loss. 

\subsection{Validity of the Convolution Model}
\label{sec:mismatch}
\begin{figure}[t!]
	\center
    \includegraphics[width=0.95\linewidth]{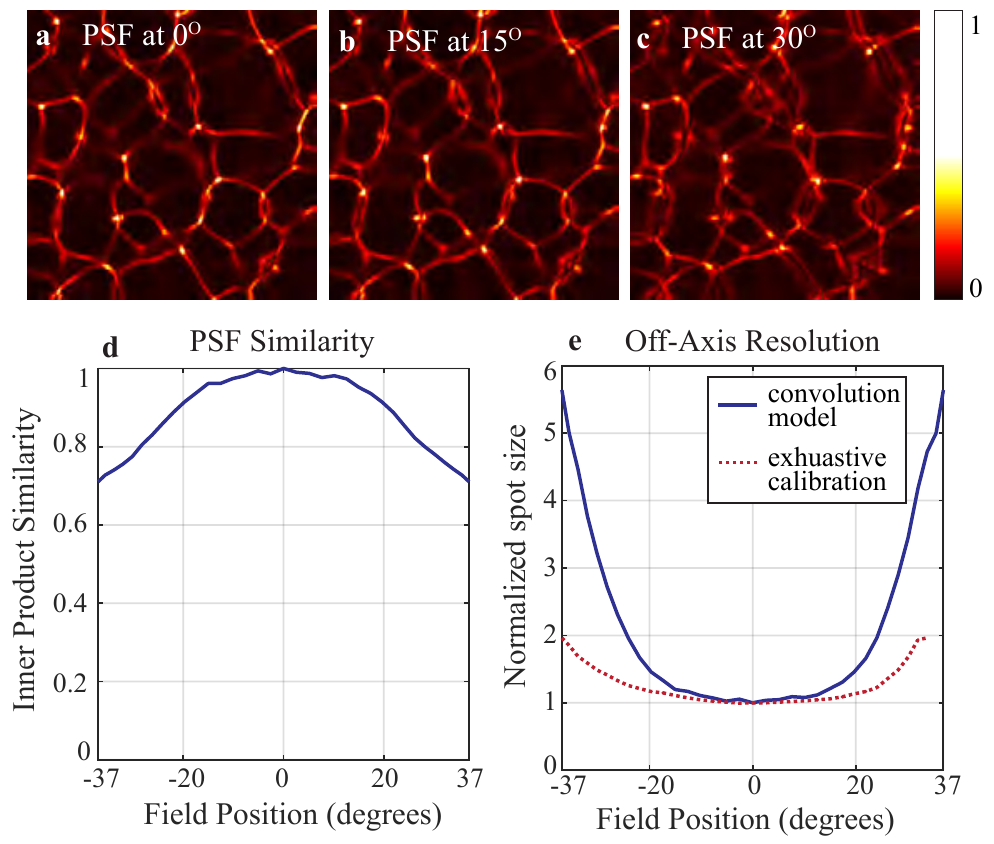}
	\caption{Experimental validation of the convolution model. (a)-(c) Close-ups of registered experimental PSFs for sources at $0^\circ$, $15^\circ$ and $30^\circ$. The PSF at $15^\circ$ is visually similar to that on-axis, while the PSF at $30^\circ$ has subtle differences. (d) Inner product between the on-axis PSF and registered off-axis PSFs as a function of source position. (e) Resulting spot size (normalized by on-axis spot). The convolution model holds well up to $\pm 15^\circ$, beyond which resolution degrades (solid). Exhaustive calibration would improve the resolution (dashed), at the expense of complexity in computation and calibration. }
	\label{fig:mismatch}
\end{figure}

In Section \ref{methods}\ref{compRecon}, we modeled the caustic pattern as shift invariant at every depth, leading to simple calibration and efficient computation. Since our convolution model is an approximation, we wish to quantify its applicability. Figure~\ref{fig:mismatch}a-c shows registered close-ups of experimentally measured PSFs from plane waves incident at $0^\circ$, $15^\circ$ and $30^\circ$. The convolution model assumes that these registered PSFs are all exactly the same, though, in reality, they have subtle differences. To quantify the similarity across the FoV, we plot the inner product between each off-axis PSF and the on-axis PSF (see Fig.~\ref{fig:mismatch}d). The inner product is greater than 75\% across the entire FoV and particularly good within $\pm 15^\circ$ of the optical axis, indicating that the convolution model holds relatively well. 

To investigate how the spatial variance of the PSF impacts system performance, we use the peak width of the cross-correlation between the on-axis and off-axis PSFs to approximate the spot size off-axis. Figure~\ref{fig:mismatch}e (solid) shows that we retain the on-axis resolution up to $\pm15^\circ$. Beyond that, the resolution gradually degrades. To avoid model mismatch, one could replace the convolution model with exhaustive calibration over all positions in the FoV. This procedure would yield higher resolution at the edges of the FoV, as shown by the dashed line in Fig.~\ref{fig:mismatch}e. The gap between these lines is what we sacrifice in resolution by using the convolution model. However, in return, we gain simplified calibration and efficient computation, which makes the large-scale problem feasible.


\section{Experimental Results} \label{sec:results}
\begin{figure*}[htbt]
	\center
    \includegraphics[width=\linewidth]{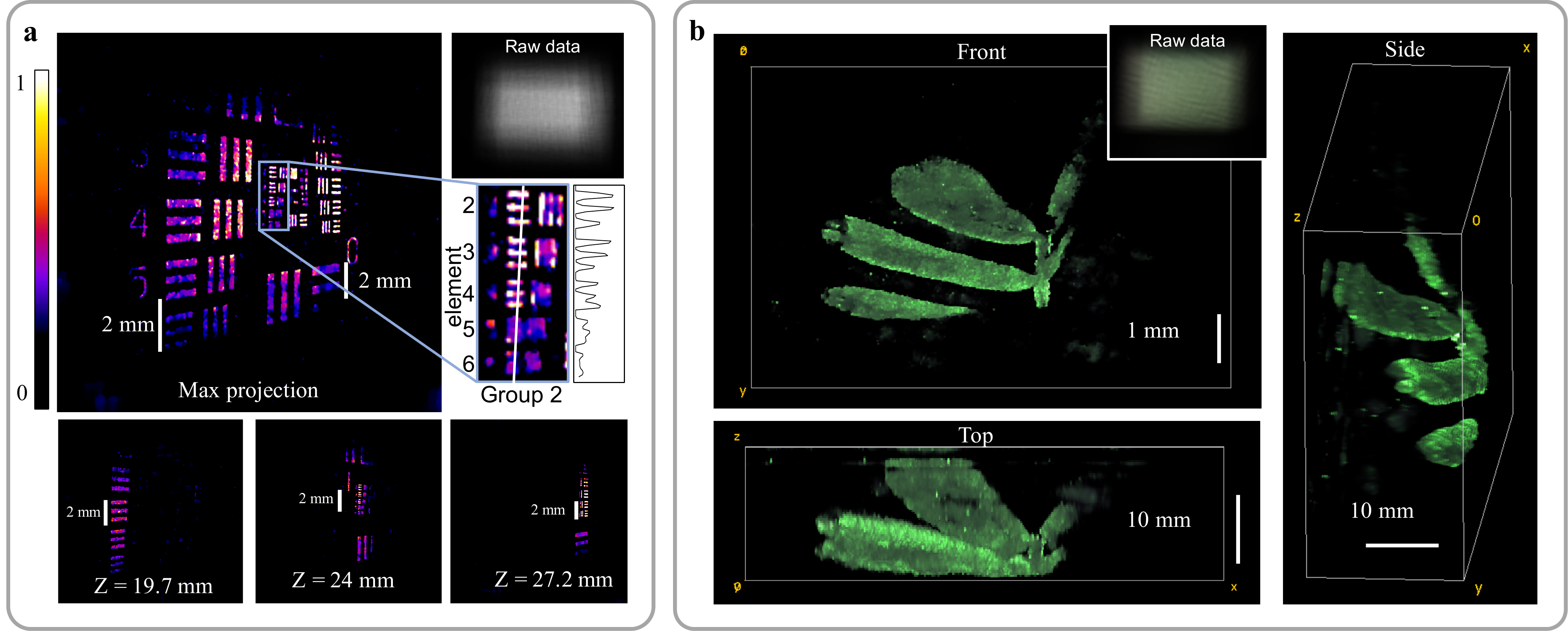}
	\caption{Experimental 3D reconstructions. (a) Tilted resolution target, which was reconstructed on a 4.2 MP lateral grid with 128 $z$-planes and cropped to 640$\times$640$\times$50 voxels. The large panel shows the max projection over $z$. Note that the spatial scale is not isotropic. Inset is a magnification of group 2 with an intensity cutline, showing that we resolve element 5 at a distance of 24 mm, which corresponds to a feature size of 79 $\mu m$ (approximately twice the lateral voxel size of $35 \mu m$ at this depth). The degraded resolution matches our 16-point distinguishability (75 $\mu m$ at 20 mm depth). Lower panels show depth slices from the recovered volume. (b) Reconstruction of a small plant, cropped to 480$\times$320$\times$128 voxels, rendered from multiple angles.}	
	\label{fig:usaf_results}
\end{figure*}

Images of two objects are presented in Fig.~\ref{fig:usaf_results}. Both were illuminated using white LEDs and reconstructed with a 3D TV regularizer. We choose a reconstruction grid that approximately Nyquist samples the two-point resolution (by $2\times2$ binning the sensor pixels to yield a 1.3 megapixel measurement). Calibration images are taken at 128 different $z$-planes, ranging from $z=10.86 mm$ to $z=36.26 mm$ (from the diffuser), with spacing set according to conditions outlined in Sec. \ref{analysis}\ref{resolution}. The 3D images are reconstructed on a 2048$\times$2048$\times$128 grid, but the angular FoV restricts the usable portion of this grid to the center 100 million voxels. Note that the resolvable feature size on this reconstruction grid can still vary based on object complexity.

The first object is a negative USAF 1951 fluorescence test target, tilted $45^\circ$ about the $y$-axis (Fig.~\ref{fig:usaf_results}a). Slices of the reconstructed volume at different $z$ planes are shown to highlight the system's depth sectioning capabilities. As described in Sec. \ref{analysis}\ref{resolution}, the spatial scale changes with depth. Analyzing the resolution in the vertical direction (Fig.~\ref{fig:usaf_results}a inset), we can easily resolve group 2 element 4 and barely resolve group 2 element 5 at $z = 24$ mm. This corresponds to resolving features $79\mu$m apart on the resolution target. This resolution is significantly worse than the two-point resolution at this depth ($50\mu$m), but similar to the 16-point resolution ($75\mu$m). Hence, we reinforce our claim that two-point resolution is a misleading metric for computational cameras, but multi-point distinguishability can be extended to more complex objects.

Finally, we demonstrate the ability of DiffuserCam to image natural objects by reconstructing a small plant (Fig.~\ref{fig:usaf_results}b). Multiple angles are rendered to demonstrate the ability to capture the 3D structure of the leaves.


\section{Conclusion}
We demonstrated a simple optical system, with only a diffuser in front of a sensor, that is capable of single-shot 3D imaging. The diffuser encodes the 3D location of point sources in caustic patterns, which allow us to apply compressed sensing to reconstruct more voxels than we have measurements. By using a convolution model that assumes that the caustic pattern is shift invariant at every depth, we developed an efficient ADMM algorithm for image recovery and simple calibration scheme. We characterized the FoV and two-point resolution of our system, and showed how resolution varies with object complexity. This motivated the introduction of a new condition number analysis, which we used to analyze how inverse problem conditioning changes with object complexity. 

\section*{Acknowledgments}

This work was funded by the Gordon and Betty Moore Foundation Grant GBMF4562 and a NSF CAREER grant to L.W. B.M. acknowledges funding from the Hertz Foundation and G. K. is a National Defense Science and Engineering Graduate Fellow. R.H. and E.B. acknowledge funding from the Swiss NSF (grants P2EZP2 159065 and P2ELP2 172278, respectively). This research was developed with funding from the Defense Advanced Research Projects Agency (DARPA), Contract No. N66001-17-C-4015. The views, opinions and/or findings expressed are those of the author and should not be interpreted as representing the official views or policies of the Department of Defense or the U.S. Government. The authors thank Dr. Eric Jonas and the Rice FlatCam team for helpful discussions.

\bibliographystyle{osajnl}
\bibliography{DiffuserCam}


\clearpage

\end{document}